\documentclass[conference]{IEEEtran}
\IEEEoverridecommandlockouts
\usepackage{cite}
\usepackage{setspace}
\usepackage{xcolor} 
\usepackage{comment}
\usepackage{hyperref}
\usepackage{amsmath,amssymb,amsfonts}
\usepackage{algorithmic}
\usepackage{graphicx}
\usepackage{subfig}
\usepackage{textcomp}
\usepackage{xcolor}
\usepackage{balance}
\usepackage{flushend}
\def\BibTeX{{\rm B\kern-.05em{\sc i\kern-.025em b}\kern-.08em
    T\kern-.1667em\lower.7ex\hbox{E}\kern-.125emX}}

\begin{document}

\title{Predictive Insights into LGBTQ+ Minority Stress: A Transductive Exploration of Social Media Discourse\\
}

\author{
\IEEEauthorblockN{Santosh Chapagain}
\IEEEauthorblockA{\textit{Utah State University} \\
santosh.chapagain@usu.edu}
\and
\IEEEauthorblockN{Yuxuan Zhao}
\IEEEauthorblockA{\textit{New Mexico State University} \\
zyx8010@nmsu.edu}
\and
\IEEEauthorblockN{Taylor K. Rohleen}
\IEEEauthorblockA{\textit{University of Florida} \\
trohleen@ufl.edu}
\and
\IEEEauthorblockN{Shah Muhammad Hamdi}
\IEEEauthorblockA{\textit{Utah State University} \\
s.hamdi@usu.edu}
\and

\IEEEauthorblockN{Soukaina Filali Boubrahimi}
\IEEEauthorblockA{\textit{Utah State University} \\
soukaina.boubrahimi@usu.edu}
\and

\IEEEauthorblockN{Ryan E. Flinn}
\IEEEauthorblockA{\textit{University of North Dakota} \\
ryanelliottflinn@gmail.com}
\and

\IEEEauthorblockN{Emily M. Lund}
\IEEEauthorblockA{\textit{University of Alabama} \\
 \textit{Ewha Womans University} \\
emlund@ua.edu}
\and

\IEEEauthorblockN{Dannie Klooster}
\IEEEauthorblockA{\textit{Oklahoma State University} \\
dannie.klooster@okstate.edu}
\and

\IEEEauthorblockN{Jillian R. Scheer}
\IEEEauthorblockA{\textit{Syracuse University} \\
jrscheer@syr.edu}
\and
\IEEEauthorblockN{Cory J. Cascalheira}
\IEEEauthorblockA{\textit{New Mexico State University} \\
cjcascalheira@gmail.com}

}

\maketitle

\begin{abstract}
Individuals who identify as sexual and gender minorities, including lesbian, gay, bisexual, transgender, queer, and others (LGBTQ+) are more likely to experience poorer health than their heterosexual and cisgender counterparts. One primary source that drives these health disparities is minority stress (i.e., chronic and social stressors unique to LGBTQ+ communities' experiences adapting to the dominant culture). This stress is frequently expressed in LGBTQ+ users' posts on social media platforms. However, these expressions are not just straightforward manifestations of minority stress. They involve linguistic complexity (e.g., idiom/lexical diversity), rendering them challenging for many traditional natural language processing methods to detect. In this work, we designed a hybrid model using Graph Neural Networks (GNN) and Bidirectional Encoder Representations from Transformers (BERT), a pre-trained deep language model to improve the classification performance of minority stress detection. We experimented with our model on a benchmark social media dataset for minority stress detection (LGBTQ+ MiSSoM+). The dataset is comprised of 5,789 human-annotated Reddit posts from LGBTQ+ subreddits. Our approach enables the extraction of hidden linguistic nuances through pretraining on a vast amount of raw data, while also engaging in transductive learning to jointly develop representations for both labeled training data and unlabeled test data. The RoBERTa-GCN model achieved an accuracy of 0.86 and an F1 score of 0.86, surpassing the performance of other baseline models in predicting LGBTQ+ minority stress. Improved prediction of minority stress expressions on social media could lead to digital health interventions to improve the wellbeing of LGBTQ+ people—a community with high rates of stress-sensitive health problems. 
\end{abstract}


\section{Introduction}
Lesbian, gay, bisexual, transgender, queer, and other sexual and gender minority (LGBTQ+) individuals experience significantly poorer physical and mental health outcomes in comparison to heterosexual and cisgender populations: higher incidences of asthma, activity limitation, cardiovascular risk \cite{c9}, human immunodeficiency virus (HIV), chronic health conditions \cite{c1}, and depression, anxiety, post-traumatic stress disorder (PTSD), substance use, self-harm, and suicidality \cite{c2, c3, c4}. Robust evidence demonstrates that social determinants of health—particularly, minority stress— contribute to the prevailing health disparities between heterosexual/cisgender and LGBTQ+ individuals \cite{c3, c8}. 

Minority stress \cite{c3} theory posits that LGBTQ+ individuals face widespread environmental threats globally \cite{c33,c34}. A main rationale for minority stress theory is that, by identifying as a sexual and gender minority, LGBTQ+ individuals face criticism and discrimination by members of the dominant culture. In an effort to adapt and to assimilate to heteronormative and cissexist societal expectations, chronic, social stressors tend to emerge. These include LGBTQ+ individuals experiencing the effects of stigmatizing and discriminating social systems, including prejudiced events (i.e., acute or chronic external stressful events), identity concealment (i.e., hiding LGBTQ+ identities due to fear of harm), expected rejection (i.e., expectations accompanied with feelings of vigilance towards possible prejudiced events), and internalized stigma (i.e., making negative societal attitudes part of their nature). Moreover, transgender and non-binary subgroups tend to present with an additional indicator of minority stress, gender dysphoria (i.e., experience of stress or discomfort stemmed from misalignment between perceived gender identity and assigned sex at birth) \cite{y34}. Despite established empirical support for the substantial health disparities faced by LGBTQ+ individuals and the central role of minority stress experiences in driving these inequities, there remains a gap in computational support for validating minority stress.

Existing applications of artificial intelligence (AI) in social science and healthcare have demonstrated their value in promoting general well-being across a multitude of disciplines\textemdash most notably, the expansion of existing theoretical frameworks in the social sciences and the emersion of AI-powered interventions \cite{y35, c23}. Natural language processing (NLP), specifically, holds the capacity to foster equity for minoritized communities \cite{y36}. That is, NLP allows for an accessible and timely detection of mental stress, opening novel pathways for the delivery of unique interventions that target minoritized individuals’ nuanced experiences and, in turn, promote the well-being of minoritized communities \cite{c23, y38}. 

For LGBTQ+ communities, the Internet—specifically, social media platforms like Reddit, on which one can post anonymously—has become a crucial space for LGBTQ+ individuals to come out, connect with peers, and seek support with less risk of experiencing disapproval and prejudice from their peers online, due to subgroup users sharing similar identities \cite{c10,c11} (e.g., LGBTQ+ users can find Reddit forums comprised of other LGBTQ+ users). Because LGBTQ+-specific Reddit forums are available for anonymous posting by LGBTQ+ people, Reddit provides an excellent space to study minority stress experiences among LGBTQ+ people. However, the expression of minority stress on these platforms is linguistically sophisticated and dynamic, which proves to be an obstacle for developing AI algorithms to study minority stress \cite{c12}. Cascalheira et al. proposed that linguistic sophistication of expressions of minority stress is threefold: (1) minority stress entails LGBTQ+\textemdash specific semantics and pragmatics, (2) psycholinguistic permutations unique to the community, and (3) lexical density (i.e., a lot of words needed to convey minority stress) \cite{c1}. Additionally, the intersecting identities of LGBTQ+ individuals (e.g., race/ethnicity, disability status, age, geographical location, etc.) inform the development and usage of novel cultural idioms that, too, convey minority stress \cite{c23}.  

Since expressions of minority stress maintain such linguistic sophistication, capturing minority stress is a great challenge for traditional NLP methods. Recently, large-scale pretraining models have demonstrated effectiveness across a spectrum of NLP tasks \cite{c13, c14}. These pretraining models, developed through extensive training on vast unlabeled corpora in an unsupervised fashion, exhibit a capacity to grasp nuanced semantic structures within the text at a considerable scale. While their intrinsic advantages for transductive learning are evident, it is noteworthy that current models designed for transductive text classification \cite{c15, c16} often disregard the integration of large-scale pretraining techniques. Consequently, the potential impact of large-scale pretraining on the classification of minority stress remains uncertain within the current landscape of transductive text classification models.

Our research aims to leverage both a transductive approach and large-scale pretraining by simultaneously training Bidirectional Encoder Representations from Transformers (BERT) and graph convolutional networks (GCN) modules. This novel approach to classifying minority stress from social media posts remains largely unexplored, and the study contributes in the following manners:
\begin{enumerate} 
\item We have designed a hybrid deep learning model, specifically BERT-GCN and RoBERTa-GCN, which leverage large-scale pretraining and graph-based representation learning for understanding relationships between different samples, enhancing predictions related to health disparities.
\item The paper aims to enhance predictive capabilities by using a language model that is based on the transformer architecture and graph neural networks. This integration allows the model to capture relationships within the entire corpus, providing a more comprehensive understanding of the data and improving predictions related to health disparities.
\item We provide a comprehensive benchmarking of various machine learning and deep learning models on the LGBTQ+ MiSSoM+ dataset \cite{c28}. It highlights the performance superiority of BERT-based architectures, particularly RoBERTa-GCN (F1 = 0.86), in predicting minority stress labels, offering valuable insights for future research in this domain.
\end{enumerate}

\section{Related Work}
Social media serves as a vital tool for LGBTQ+ identity exploration and expression, as it allows individuals a platform to present varied facets of themselves across different networks and to ascertain support for their LGBTQ+ identity \cite{c18,c19}. While this study focuses on text data analysis to contribute to existing research \cite{c20,c21,c22,c23}, the broader potential lies in employing a transductive approach to analyze LGBTQ+ users’ social media posts in media formats other than text (e.g., short-form and long-form video, voice recording, multimedia) to reduce LGBTQ+ health disparities. The multifaceted nature of social media content (text, image, video) suggests an opportunity for innovation in modeling social determinants beyond textual information. Our results can provide immediate value to applied social scientists for interventions in LGBTQ+ health, with the potential to extend into a multimodal prediction of minority stress and related health constructs using non-text data—a promising yet underexamined area in applied big data analytics \cite{c20}. Importantly, social media emerges as a unique data source for LGBTQ+ research, as it provides exceptional reliability for studying psychological processes, such as the establishment of belonging and community \cite{c24}. Prior studies have demonstrated the effectiveness of deep learning models, such as Bidirectional Encoder Representations from Transformers (BERT) \cite{c13}, recurrent neural networks (RNNs) \cite{c27} with its variants like bidirectional long short-term memory (Bi-LSTM), bidirectional gated recurrent unit (BiGRU), convolutional neural networks (CNNs), and hybrid models, in predicting social determinants of health disparities, including cyberbullying and racism on Twitter \cite{c25}. Multichannel CNN achieved a notable F1 score of 0.97 in classifying expressions of general stress on the Twitter dataset, which consists of interview responses from 38 students and stress-related tweets with certain hashtags. \cite{c25}. Hybrid models, such as BERT-LSTM, have shown promise in improving the prediction of misogynistic speech on Twitter (F1 = 0.81) \cite{c26}.

Past attempts to classify minority stress on Reddit users’ text data \cite{c23} achieved an F1 score of 0.75 with models such as logistic regression. However, their limited ability to handle sequential data led to drawbacks. Cascalheira et al. \cite{c21} introduced a Bi-LSTM but achieved only an F1 score of 0.61, indicating architectural limitations. Later, the combination of BERT-CNN exhibited excellent performance for both composite minority stress and factors of minority stress \cite{c23}. While these models have successfully predicted health disparities related to external discrimination, studies focusing on composite minority stress have shown room for improvement.

In contrast to the previous research, our purposed method aims to enhance predictive capabilities by leveraging large-scale pretraining (BERT, RoBERTa) and neural networks. It employs graph neural networks (GNNs) to model relationships between labeled and unlabeled documents, utilizing the similarity between them within the whole corpus. This approach extends beyond previous models to explore the potential of neural networks in understanding the relationships between different samples and enhancing predictions related to health disparities.

\section{Notations and Preliminaries}
\subsection{Graph Definition}
Formally, in a graph with $n$ nodes, $G = (V, E)$ is characterized by a set of $V = \{v_1, v_2, v_3, \dots, v_n\}$ and a set of edges $E = \{e_{ij}\}_{i,j=1}^{n}$ denoting relationships between the nodes. The adjacency matrix $A$, with dimensions $n \times n$, serves as a fundamental representation of the graph. In unweighted graphs, $A_{ij}$ equals $1$ if an edge exists between nodes $i$ and $j$, and $0$ otherwise. Conversely, in weighted graphs, $A_{ij}$ represents the weight of the relationship between nodes $i$ and $j$, with $0$ indicating no relationship. For undirected graphs, $A$ is symmetric ($A_{ij} = A_{ji}$), while for directed graphs, it is asymmetric ($A_{ij} \neq A_{ji}$). In this work, we constructed an undirected heterogeneous graph for the whole corpus. There are two node types (documents and words), and the edges are formed using term frequency-inverse document frequency (TF-IDF) \cite{c29} for word-document connections and positive point-wise mutual information (PPMI) for word-word connections.

\subsection{Graph Convolutional Networks (GCN)}
Graph Convolutional Networks (GCN) \cite{kipf2016semi}, are neural networks that operates through neighborhood aggregation and message passing mechanism. GCNs generate embedding vectors of nodes based on the properties of their neighbor nodes. Additionally, a feature matrix $X \in \mathbb{R}^{n \times m}$ containing $n$ nodes with $m$-dimensional feature vectors is defined, with each row $x_v \in \mathbb{R}^{m}$ representing the feature vector for node $v$. An adjacency matrix $A$ and its associated degree matrix $D$, where $D_{ii} = \sum_{j} A_{ij}$, are introduced.

GCN processes information solely from immediate neighbors in one convolutional layer. By stacking multiple GCN layers, it integrates information from multi-hop neighborhoods. For a single-layer GCN, the new $k$-dimensional node feature matrix $L^{(1)} \in \mathbb{R}^{n \times k}$ is computed as

\begin{equation}
L^{(1)} = \sigma(\tilde{A}XW^{(0)})
\end{equation}

where $\tilde{A} = D^{-1/2}AD^{-1/2}$ is the normalized symmetric adjacency matrix, and $W^{(0)} \in \mathbb{R}^{m \times k}$ is a weight matrix. Here, $\sigma$ represents an activation function such as ReLU ($\sigma(x) = \max(0, x)$). Stacking multiple GCN layers allows the incorporation of higher-order neighborhood information: 

\begin{equation}
L^{(i)} = \sigma(\tilde{A}L^{(i-1)}W^{(i)})
\end{equation}

where $i$ denotes the layer number, and $L^{(0)} = X$.

\section{Methods}
 
\subsection{Neural Network Architectures}
The BERT-GCN architecture, shown in Figure 1, utilizes a BERT model to initialize representations for document nodes in a heterogeneous Reddit corpus graph. These initialized representations serve as inputs to a Graph Convolutional Network (GCN), where iterative updates based on the graph structures take place. The resulting outputs are treated as final representations for document nodes, which are forwarded to a linear layer and a softmax layer for predictions. Below are detailed descriptions and the roles of each component involved in this architecture.

\begin{figure*}[!htbp]
\centerline{\includegraphics[width=1\textwidth]{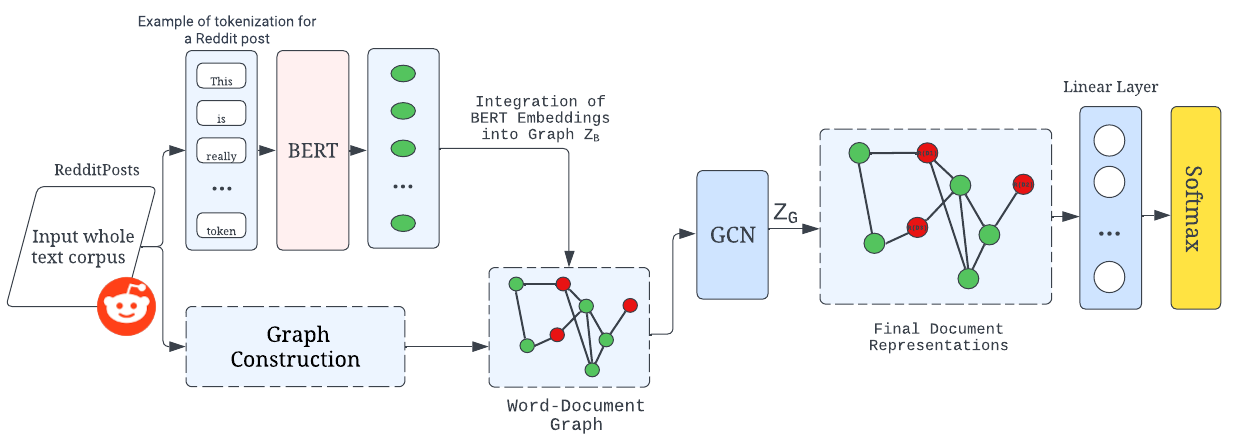}}
\caption{BERT-GCN Network Architecture in Action.}
\label{architecture1}
\end{figure*}

\subsubsection{BERT} BERT is built upon the transformer architecture, utilizing bidirectional transformer encoder layers to learn contextual word relationships. Each transformer encoder block comprises two sublayers: a multi-head self-attention mechanism and a position-wise fully connected feed-forward network. Residual connections are applied within each sub-layer. The architecture of the BERT transformer encoder block is shown in Figure 2. During pre-training, BERT utilizes masked language model-based and next-sentence prediction objectives \cite{c23}. Fine-tuning for specific tasks is achieved by adding task-specific layers to the pre-trained BERT model.

\begin{figure} 
\centerline{\includegraphics[width=0.40\linewidth]{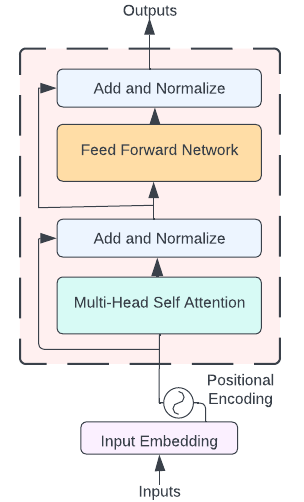}}
\caption{The BERT layer.}
\label{architecture2}
\end{figure}

We represent the number of document nodes by $N_{doc}$, and the number of word nodes by $N_{word}$ that includes both training and test sets. The document node embeddings, Z$_{doc} \in \mathbb{R}^{N_{doc} \times d}$, where $d$ is the dimension of the embedding, and the initial node feature matrix is given by equation (3): 

\begin{equation}
    X = \begin{pmatrix}
    Z_{doc} \\
    0
    \end{pmatrix}_{(N_{doc}+N_{word}) \times d}
\end{equation}

Now, the X is fed into a dense layer with a softmax, and the output, $Z_B$ is expressed as below:
\begin{equation}
    Z_B = softmax(W X)
\end{equation}
where $W$ is the weight matrix responsible for transforming the node features.

\subsubsection{Graph Construction}

We constructed a heterogeneous graph following TextGCN \cite{c15}, and BertGCN \cite{c17} by using TF-IDF for word-document edges and PPMI for word-word edges, we define the edge weights between nodes i and j with equation (5). Figure 3 illustrates the top five words from ten documents using the same equation to create a heterogeneous graph.

\begin{equation}
    A_{i,j} = 
    \begin{cases}
        \text{TF-IDF}(i, j), & \text{if } i \text{ is a document, } j \text{ is a word} \\ 
        \text{PPMI}(i, j), & \text{if } i, j \text{ are words and } i \neq j \\
        1, & \text{if } i = j \\
        0, & \text{otherwise}
    \end{cases}
\end{equation}

The Positive Pointwise Mutual Information (PPMI) value of a word pair \( (i, j) \) is calculated as:
\begin{equation}
\text{PPMI}(i, j) = \log \frac{p(i, j)}{p(i)p(j)} 
\end{equation}

Where:
\begin{itemize}
    \item \( p(i, j) = \frac{\text{\#}W(i, j)}{\text{\#}W} \) represents the probability of occurrence of the word pair \( (i, j) \) in the corpus.
    \item \( p(i) = \frac{\text{\#}W(i)}{\text{\#}W} \) is the probability of occurrence of word \( i \) in the corpus.
    \item \( \text{\#}W(i) \) denotes the number of sliding windows in the corpus that contain the word \( i \).
    \item \( \text{\#}W(i, j) \) denotes the number of sliding windows that contain both words \( i \) and \( j \).
    \item \( \text{\#}W \) is the total number of sliding windows in the corpus.
\end{itemize}

A positive PPMI value indicates a high semantic correlation between words in the corpus, while a negative PPMI value suggests little to no semantic correlation. Consequently, edges are only added between word pairs with positive PPMI values.

After constructing the graph, X is fed into GCN layers, and the feature matrix of the i-th layer is given by equation (2). The output of GCN is then fed to the softmax layer as is given by equation (7):
\begin{equation}
    Z_G = softmax(gcn(X, \widetilde{A}))
\end{equation}

where $gcn$ is GCN model as shown in the equation (1) and (2).

\begin{figure}[htbp]
\centerline{\includegraphics[width=1\linewidth]{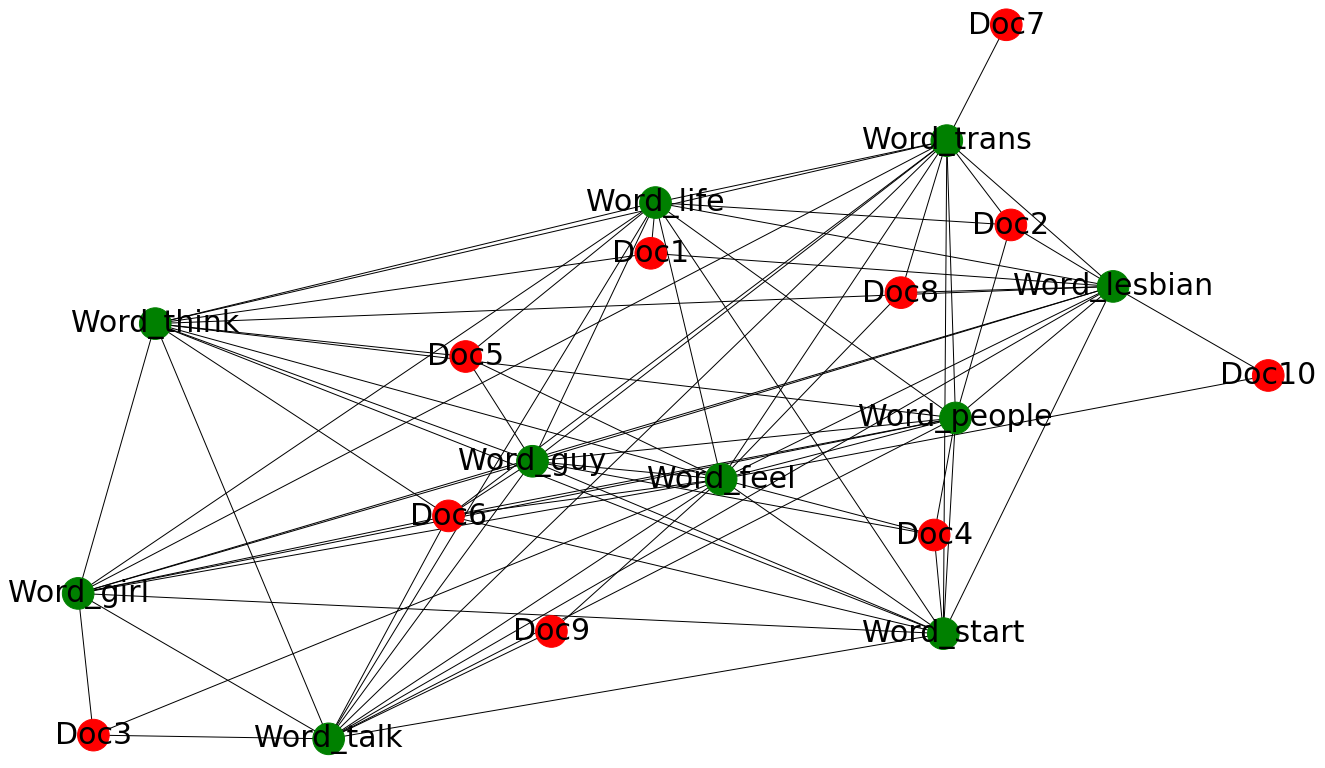}}
\caption{Heterogeneous Graph: TF-IDF and PPMI Edge Construction from 10 Documents with Top 5 Words of LGBTQ+ MiSSoM+ dataset.}
\label{graphArch}
\end{figure}

\subsubsection{BERT-GCN}
BERT-GCN performs the linear interpolation of the prediction from BERT and GCN model, and the final result is shown in equation (8): 

\begin{equation}
    Z_{Final} = \lambda Z_G + (1 - \lambda) Z_B
\end{equation}

where $\lambda\in (0,1) $ is the hyper-parameter that controls two models. When $\lambda = 0$, we fully use the BERT module, and when $\lambda = 1$, we use the GCN component fully. This allows us to balance between two models to achieve the best result for our classification task.

\section{Experiments}
\subsection{Dataset}

We used the LGBTQ+ Minority Stress on Social Media advanced version (MiSSoM+) dataset \cite{c28}. The dataset is composed of unprocessed text sourced from the social media platform Reddit, spanning a period from 2012 to 2021. The creators of the LGBTQ+ MiSSoM+ dataset gathered extensive data from seven prominent LGBTQ+-themed subreddits, namely \textit{r/actuallesbians},\textit{ r/ainbow},\textit{ r/bisexual}, \textit{r/gay}, \textit{r/genderqueer}, \textit{r/questioning}, and \textit{r/trans}. A total of 27,796 posts were collected, with LGBTQ+ health experts annotating 5,789 of them, resulting in the creation of the LGBTQ+ MiSSoM+ dataset. The dataset comprises 4,551 positive class labels and 1,238 negative class labels. The annotating team was mostly comprised of members of LGBTQ+ subgroups and of researchers with diverse social identities (e.g., racial/ethnic identities, sexual and gender identities, age, etc.), fields of expertise, and levels of education. With the anonymous and publicly accessible nature of Reddit posts, the collection and annotation process for the MiSSoM+ dataset was exempted from Institutional Review Board (IRB) review. Figure 4 shows the word clouds of positive (i.e., presence of minority stress) and negative (i.e., absence of minority stress) labels of the dataset. The most frequently occurring words in positive-labeled posts are “feel”, “guy”, “feeling”, “friend”, “men”, and 'gender'. Similarly, in negative-labeled posts, words such as “thank”, “friend”, “girl”, “guy”, and “men” are frequently mentioned. Additionally, Table 1 displays text examples of minority stress labels in the LGBTQ+ dataset, paraphrased to protect the anonymity of Reddit users.

\begin{figure}[!htb]
    \centering
    \subfloat[\centering Positive]{{\includegraphics[width=7cm]{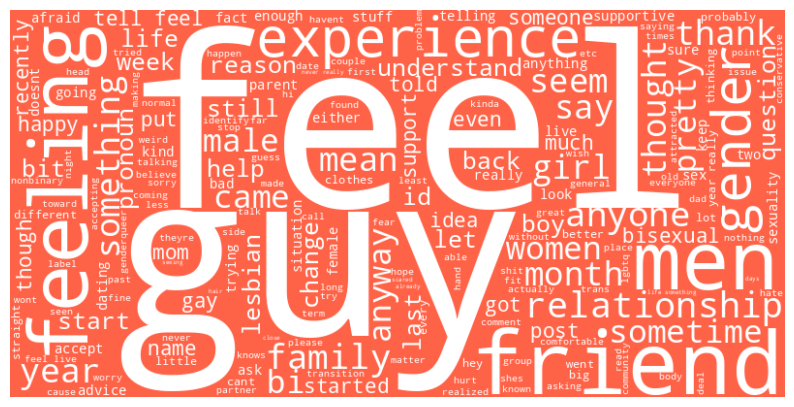} }}%
    \qquad
    \subfloat[\centering Negative]{{\includegraphics[width=7cm]{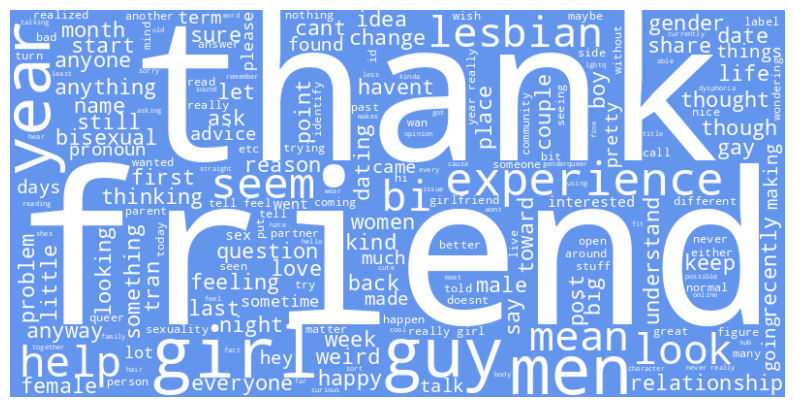} }}%
    \caption{WordCloud on LGBTQ+ MiSSoM+ dataset labels.}%
    \label{fig:example2}%
\end{figure}

\begin{table}[htbp]
  \centering
  \caption{Text examples on LGBTQ+ MiSSoM+ dataset.}
  \begin{tabular}{p{2.5 cm} p{4.6cm}} 
  \hline & \\[-1.5ex]
    Label &  Text \\
  \hline
    0 (minority stress absent) & I am cisgender woman. I have lived with my partner for years but I have always been attracted to other women. But I'm mostly attracted to their aesthetics not sexually. \\
    1 (minority stress present) & I connect being straight to a successful future. I have to be straight if I want things in life.  Being a lesbian will mean having a life where everything I want will be extremely hard to get. \\
  \hline
  \end{tabular}
\end{table}

The corpus is then split into a 70:30 ratio for training and testing, and 10\% of the training dataset is reserved for validation, as demonstrated in \cite{c15}. During the graph construction, the entire corpus is utilized to generate a heterogeneous graph, resulting in a total of 54,751 nodes and 7,145,345 edges consisting of both the training and test sets. 

\subsection{Baseline Algorithms}
We compared the classification performance with several machine learning and deep learning models including hybrid models.

\begin{itemize}

\item{Naïve Bayes:} Naive Bayes calculates the probability of a data point belonging to a class and predicts the class with the highest probability; the baseline used Multinomial Naive Bayes with TF-IDF \cite{c29} representation for text feature extraction.

\item{Logistic Regression:} Logistic regression is a classification algorithm that uses the logistic loss function, and is trained with TD-IDF and L2 regularization \cite{c30}.

\item{Support Vector Machine (SVM):} SVM maximizes the margin between support vectors, aiming to separate data points into classes with minimal error and enhanced generalization \cite{c30}. Trained with TF-IDF, it employs a regularization parameter of 1 and the radial basis function kernel.

\item{Random Forest:} An ensemble learning method that combines 100 decision trees trained on TF-IDF features, making predictions through a majority vote or averaging individual tree predictions \cite{c30}.

\item{AdaBoost:} An ensemble learning algorithm that iteratively trains weak classifiers on weighted versions of the data, assigning higher weights to misclassified samples in each iteration to focus subsequent classifiers on those instances \cite{c32}. The baseline used a decision tree classifier with a max depth of 1, 50 estimators, and TF-IDF features.

\item{Multi-Layer Perceptron (MLP):} A feed-forward neural network with multiple layers of fully connected neurons, trained using TF-IDF with default hyperparameters (hidden layer: 100 neurons, adam optimizer, relu activation) \cite{c30}

\item{BiLSTM:} The BiLSTM (Bidirectional Long Short-Term Memory) model represents a sophisticated form of recurrent neural network architecture specifically designed for the processing of sequential data \cite{c31}. It utilizes a two-layer BiLSTM with a hidden size of 256, with pre-trained Glove word embeddings (dimension 100), and a fully connected layer as in the paper \cite{c21}.

\item{BERT-BiGRU:} A hybrid model that combines BERT's contextual embeddings with two layers of Bidirectional Gated Recurrent Units (BiGRU). Using a hidden dimension of 256 in the BiGRU layers, the model captures intricate patterns bidirectionally in input sequences. A linear layer finalizes predictions \cite{c23}.

\item{BERT-CNN:} BERT is the embedding layer with three convolutional layers consisting of 100 filter counts on each layer and a diverse kernel size of 3, 4, and 5 on each layer \cite{c23}.
\item{BERT:} A pre-trained natural language processing model that captures contextual relationships in words \cite{c13}. The baseline model configuration includes setting the batch size to 32, initializing BERT with \textit{bert-large-uncased}, with a learning rate of 1e-05 and dropout of 0.5. 
\item{RoBERTa:} 
A robustly optimized BERT-based model, designed to enhance NLP tasks such as question answering, text classification, and language modeling \cite{c14}. The baseline configuration shares the same settings as BERT for most parameters.

\end{itemize}

\subsection{Experimental Settings}
We conducted all our experiments on the Linux server. The server featured dual Intel Xeon Gold 5220R processors, each comprising 24 cores clocked at 2.20 GHz, with a substantial 35.75 MB cache. Additionally, the server was equipped with four NVIDIA RTX A5000 GPUs, having 24 GB of VRAM per GPU. The training was conducted using PyTorch 1.13.0 with CUDA 11.1 to implement BERT-GCN. We opted for a two-layer GCN as it demonstrated superior performance compared to a single layer. For BERT-GCN and RoBERTa-GCN, we utilized the \textit{bert-large-uncased} and \textit{roberta-large}, respectively, as they outperformed \textit{bert-base-uncased } and \textit{roberta-base}. The learning rate for the GCN was set to 1e-3, while the BERT module used a learning rate of 1e-5. GAT variants were also trained with the same hyperparameters, and the attention head count was set to 8. Both the BERT-GCN and BERT-GAT were trained for 50 epochs. The source code can be found on GitHub.\footnote{\url{https://github.com/chapagaisa/transductive}}.

\subsection{Performance Measures}
We used accuracy and weighted F1 score as our performance metric. Accuracy is a fundamental measure in classification tasks that quantifies the overall correctness of predictions by comparing the number of correctly classified instances to the total number of instances. On the other hand, a weighted F1 score is a metric that takes into account both precision and recall across multiple classes, offering a more detailed evaluation that provides a balanced assessment that considers both false positives and false negatives.

\subsection{Results}
Table 2 presents a performance comparison of various models on the LGBTQ+ MiSSoM+ dataset, evaluating their accuracy and F1-score for the composite minority stress. Traditional machine learning models, including Naïve Bayes, Logistic Regression, SVM (Linear), Random Forest, AdaBoost, and MLP, demonstrated competitive but varied results, with accuracy ranging from 0.75 to 0.81 and F1-scores from 0.69 to 0.79.

Moving to deep learning models, BiLSTM did not perform as well compared to others, but BERT-based architectures (BERT-BiGRU, BERT-CNN, BERT, RoBERTa) showed significant improvements, with accuracy scores exceeding 0.80 and F1-scores surpassing 0.78. RoBERTa-GCN performed the best among all other classifiers, with an accuracy of 0.8624 and an F1 score of 0.8536, demonstrating its effectiveness in capturing nuanced patterns related to minority stress labels. Compared with BERT and RoBERTa, there is a slight performance boost from BertGCN and RoBERTaGCN, suggesting that the models take advantage of the graph structure. 

Figure 5 displays the Receiver Operating Characteristic (ROC) diagram of various classifiers along with their respective Area Under Curve (AUC). The ROC-AUC of RoBERTa-GCN is 0.8735, slightly lower than that of RoBERTa and BERTCNN. While ROC-AUC is an important metric for illustrating classifier accuracy by distinguishing between positive and negative classes across diverse thresholds, the F1 score is more attuned to class imbalance. Our models outperform all classifiers in terms of F1 score, indicating their excellence in correctly identifying instances of the minority class and effectively addressing class imbalance.

\begin{figure} 
\centerline{\includegraphics[width=1\linewidth]{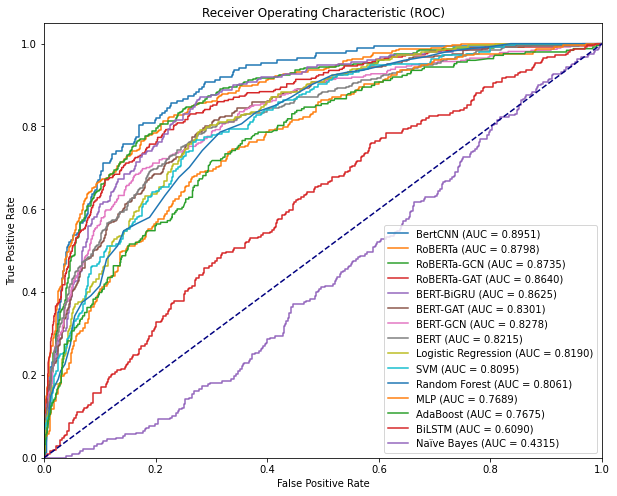}}
\caption{ROC-AUC of different classifiers.}
\label{architecture7}
\end{figure}

\begin{table}[ht]
\centering
\caption{Performances comparison on minority stress label on the LGBTQ+ MiSSoM+ dataset.}
\begin{tabular}{p{3.6cm} p{2cm} p{2cm}}
\hline 
Model & Accuracy & F1-Score \\
\hline 
Naïve Bayes & 0.7932 & 0.6926 \\
Logistic Regression  & 0.8103 & 0.7832 \\
SVM (Linear) & 0.8103 & 0.7902 \\
Random Forest & 0.7922 & 0.7018 \\
AdaBoost & 0.7902 & 0.7845 \\
MLP  & 0.8026 & 0.7865 \\
BiLSTM  & 0.7545 & 0.7236 \\
BERT-BiGRU  & 0.8432 & 0.8311 \\
BERT-CNN & 0.8608 & 0.8422 \\
BERT & 0.8112 & 0.8106 \\
RoBERTa & 0.8457 & 0.8415 \\
BERT-GCN & 0.8198 & 0.8116 \\
\textbf{RoBERTa-GCN} & \textbf{0.8624} & \textbf{0.8536} \\
BERT-GAT & 0.8140 & 0.8150\\
RoBERTa-GAT & 0.8307 & 0.8334\\
\hline
\end{tabular}
\end{table}

\subsection{Ablation Study}
$\lambda$ is the hyperparameter in BERT-GCN that controls the weight assigned to the BERT module and GCN module. Different values of $\lambda$ lead to different accuracy and F1 Score outcomes. Figure 6 illustrates the classification results on the test dataset for the LGBTQ+ MiSSoM+ dataset with different values of $\lambda$. When we set $\lambda$ to 0, we only use the BERT module. When we set $\lambda$ to 1, we only use the GCN module. We ran several experiments to get optimal values and observed that the best classification results, in terms of both accuracy and F1-score, were achieved when $\lambda = 0.2$. The performance boost with the addition of the GNN component is approximately 1.05\% for BERT in terms of accuracy and 0.12\% in terms of F1 score, while for RoBERTa, it is approximately 1.97\% for accuracy and 1.44\% for F1 score. It is commonly acknowledged that pretraining based on BERT holds greater significance compared to graph learning utilizing GNNs. However, it is imperative to note that employing GNN-based graph learning, when appropriately emphasized and integrated into the model architecture, has the potential to significantly enhance overall accuracy. 

\begin{figure}[htbp]
\centerline{\includegraphics[width=1\linewidth]{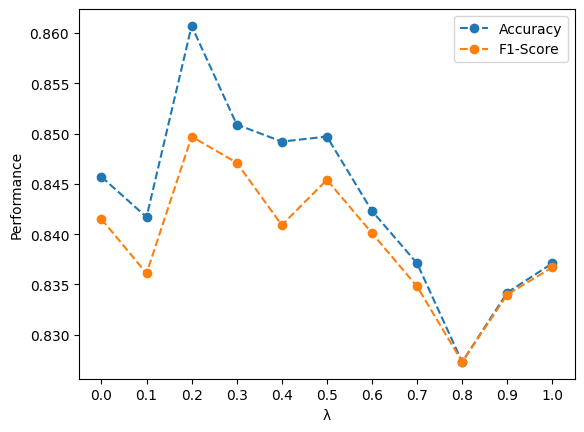}}
\caption{Test Accuracy and F1 score on LGBTQ+ MiSSoM+ dataset with varying $\lambda$ values on RoBERTa-GCN.}
\label{lambda}
\end{figure}

\section{DISCUSSION AND CONCLUSION}
This is the first study to leverage transductive modeling coupled with large-scale pretraining to predict minority stress among LGBTQ+ people who use Reddit. Fifteen different models were tested, running the gamut from traditional supervised machine learning to novel neural network architectures. The transductive-pretrained hybrid model, RoBERTa-GCN, yielded superior performance. Several ethical considerations, implications for digital health interventions, and limitations are discussed to frame our experimental results.
\subsection{Ethical Considerations}
In conducting these analyses, we are committed to following Reddit’s User Agreement \cite{y43}, the American Psychological Association's Ethical Principles of Psychologists and Code of Conduct \cite{y44}, and the Association for the Advancement of Artificial Intelligence’s Code of Professional Ethics and Conduct and Diversity Statement \cite{y45} throughout our study. We emphasize the need for careful consideration of end-use products when studying the social determinants of LGBTQ+ health inequities, as highlighted in past work \cite{c20,c21,c22,c23}.

This study’s key ethical strengths include safeguarding LGBTQ+ Reddit users’ privacy and actively involving LGBTQ+ individuals in all aspects of the research. As part of our commitment to nonmaleficence, our research team is comprised of members of the LGBTQ+ community \cite{c23}. We aim to maintain Reddit user anonymity and privacy by restricting access to the datasets, and releasing only the code for analysis reproduction due to potential safety concerns. While scraping Reddit data without explicit consent is common practice, we acknowledge the need for improved communication with LGBTQ+ social media users and AI scientists. Despite not publicly releasing the data, we recognize the enduring potential if unlikely risk of malicious replication of our modeling techniques, which could lead to the identification of LGBTQ+ users and forms of online harassment (e.g. cyberbullying). 

Additionally, we acknowledge the mindfulness and nuance necessitated in working with a vulnerable and historically marginalized community, as LGBTQ+ research maintains/elicits unique risks aside of common potential risks. An example of a unique risk is “outing” a closeted individual (i.e., revealing that they are not cisgender and/or heterosexual), as they may face mental decline, diminished safety, or legal consequences in certain regions of the world as a result. Our commitment to minimizing harm and maximizing benefits guides our pursuit of understanding the relationship between minority stress and LGBTQ+ health. 

Further, in an effort to contribute thoughtfully to the social sciences, to data science, and to the development of digital health interventions, the MiSSoM+ dataset is well-developed in its accuracy. This advantage can be attributed to the creation of an annotating team constitution and the diverse composition of the annotating team: An annotating team constitution provokes the development of precise definitions for contextualizing LGBTQ+ Reddit user experiences—definitions which were guided by their expertise in LGBTQ+ health and their own lived experiences \cite{y40,y42}. Thus, the diverse cultural backgrounds of our team members facilitate accuracy in linguistically capturing cultural nuances. Their heterogeneity offers a variety of academic disciplines and a keen adeptness in LGBTQ+ studies, both of which yield a more accurate dataset. 

The current study, due to its absence of direct interactions with human subjects and use of publicly available data, does not qualify for an ethics board revision process. Yet, in consideration of the numerous risks in working with vulnerable populations while employing a computational method, we encourage future AI and social scientists to enhance our current knowledge of confidentiality, expand on the functionality of an ethics board, and continue this commitment of systematic ethics reviews in computational social science research. 

Lastly, we acknowledge the timeliness of our models with respect to constantly evolving language in social media. Common LGBTQ+-specific semantics and pragmatics remain fluid and continue to evolve with increased understanding of LGBTQ+ identities among both the LGBTQ+ communities and the general public \cite{y41}. We suggest continuous education on sophisticated linguistics of LGBTQ+ communities and timely updates of available datasets to maintain integrity in future similar computational studies. With respect to sophistication in linguistics, we also acknowledge that the annotating team for the MiSSoM+ dataset lacks expertise or lived experiences of certain cultural communities (e.g., lack of understanding of diverse queer culture and living situations on an international level due to limited experiences with diverse geographical locations of members). We encourage development of datasets with further consideration of intersectionality and contextual factors.

\subsection{Implication for Digital Health Interventions}
The precise identification of minority stress within social media language holds immense potential for impactful innovations in the healthcare sector and social policy. Our findings have the capacity to support and instruct the development of personalized digital health interventions for LGBTQ+ individuals. These interventions might include stress-reduction smartphone apps triggering coping strategies upon detecting minority stress. Moreover, due to the uniquely challenging process of LGBTQ+ identity development, our findings intend to support the conception of identity exploration interventions.

Our models could also optimize content delivery in existing health or identity exploration apps, as well as improve safety or crisis resources through assigning AI-informed modules to a user based on detected stress levels. Additionally, the accurate detection of minority stress could prompt timely booster sessions or interventions, potentially reducing minority stress and thereby improving LGBTQ+ individuals’ mental health outcomes and, thus, safety. These applications could significantly benefit LGBTQ+ individuals’ quality of life by addressing and mitigating the impacts of minority stress, ultimately beginning to bridge the inequity of health among heterosexual/cisgender and LGBTQ+ populations.

\subsection{Limitations and Future Research}
Although our paper outperforms all the baseline models that were successful in predicting LGBTQ+ Minority Stress, several limitations point toward avenues for future work. Firstly, the use of a pre-trained BERT model without fine-tuning the embedding layer may limit the model's task-specific performance, urging researchers to explore the potential benefits of fine-tuning BERT for classification tasks. Secondly, the transductive nature of BERT-GCN hampers its agility in processing unseen test documents, indicating a need to investigate and integrate inductive models to enhance adaptability. Thirdly, the GAT module primarily focuses on the 1-degree neighborhood, prompting future work to consider expanding the attention range and incorporating sub-graph attention learning for a more comprehensive understanding. Additionally, the graph construction process, relying solely on document statistics, may be sub-optimal compared to models that can automatically establish edges between nodes. Future research should work into methods for automated edge construction, potentially offering a more robust graph structure and enhancing overall model performance. 

From the lens of social science, we applied a human-annotated dataset based on the minority stress theory. While the minority stress theory is considered a well-established theory in contextualizing social stressors LGBTQ+ individuals experience, we acknowledge that the model does not account for intersecting identities (e.g., being a gender/sexual minority while identifying with low-income social status, etc.) \cite{y46,y47}. Further research should continue expanding on minority stress with a broader consideration of social identities and their intersectionality.

\subsection{Conclusion}
In this study, we used BERT-GCN, a model that leverages the strengths of both pre-trained language models and a graph neural network for predicting LGBTQ+ minority stress. Employing a real-world social media dataset (LGBTQ+ MiSSoM+), we created a heterogeneous graph and used BERT representation for representing documents as node embeddings. We jointly trained BERT and GCN to improve our performance and the experiment shows the RoBERTa-GCN performs the overall best to all baseline models \cite{c20,c21,c22,c23}, improving the prediction of minority stress in comparison to BERT-CNN. 

\section{Acknowledgements}
Cory J. Cascalheira is supported as a RISE Fellow by the National Institutes of Health (R25GM061222). Ryan E. Flinn is supported as a Scholar/Trainee by the following training programs, each of which are funded by the National Institute on Drug Abuse (NIDA): the HIV/AIDS, Substance Abuse, and Trauma Training Program at the University of California, Los Angeles (R25DA035692); the Lifespan/Brown Criminal Justice Research Training Program on Substance Use, HIV, and Comorbidities (R25DA037190); the JEAP Initiative (R24DA051950); and the Brandeis-Harvard SPIRE Center Substance Use Disorder Systems Performance Scholars Program (P30DA035772). Shah Muhammad Hamdi is supported by the CISE and GEO directorates under NSF awards \#2301397 and \#2305781. Soukaina Filali Boubrahimi is supported by CISE and GEO Directorates under NSF awards \#2204363, \#2240022, \#2301397, and \#2305781. Emily M. Lund is a visiting professor at Ewha Women’s University, and resources purchased with Ewha funds were used in the preparation of this manuscript. Jillian R. Scheer is supported by a Mentored Scientist Development Award (K01AA028239-01A1) from the National Institute on Alcohol Abuse and Alcoholism.

\bibliography{output.bib} 
\bibliographystyle{IEEEtran}

\end{document}